\documentclass[11pt,a4paper]{article}

\usepackage[utf8]{inputenc}
\usepackage[T1]{fontenc}
\usepackage{lmodern}
\usepackage[margin=1in]{geometry}
\usepackage{authblk}
\usepackage{microtype}
\usepackage{amsmath,amssymb}
\usepackage{graphicx}
\usepackage{booktabs}
\usepackage{multirow}
\usepackage{placeins}
\usepackage{enumitem}
\usepackage{xcolor}
\usepackage[numbers,sort&compress]{natbib}
\usepackage[colorlinks=true,linkcolor=blue!50!black,citecolor=blue!50!black,urlcolor=blue!50!black]{hyperref}

\setlist[itemize]{itemsep=2pt,topsep=4pt}

\title{\bfseries Benchmarking Model Compression Techniques Across Heterogeneous Edge Platforms: A Predictive Maintenance Use Case for Rotating Machinery}

\author[1]{Matheus Calheira Guimarães de Oliveira}
\author[2]{Marco Aurélio da Silva Cruz}
\author[3,4]{Regis Cardoso}
\author[5]{Lucas Cambuim}
\author[6]{Rafael Macieira}
\author[1]{Arthur Paim}
\author[1]{Helaine Neves}
\author[1]{Alex Santos}
\author[1]{Lilian Guarieiro}
\author[4,7]{Jean Schmith}
\author[2,8,9]{Prashant Kumar}
\author[1,2,8,9]{Erick G. Sperandio Nascimento\thanks{Corresponding author: \texttt{erick.sperandio@surrey.ac.uk}}}

\affil[1]{SENAI CIMATEC University, Salvador, BA, Brazil}
\affil[2]{Surrey Institute for People-Centred AI, School of Computer Science and Electronic Engineering, Faculty of Engineering and Physical Sciences, University of Surrey, Guildford, UK}
\affil[3]{SENAI Institute of Innovation in Embedded Systems, Florianópolis, SC, Brazil}
\affil[4]{UNISINOS University, São Leopoldo, RS, Brazil}
\affil[5]{Department of Statistics and Informatics (DEINFO), Federal Rural University of Pernambuco (UFRPE), Recife, PE, Brazil}
\affil[6]{Integrated Center for Advanced Security Systems, Center for Advanced Studies and Systems of Recife (CESAR), Recife, PE, Brazil}
\affil[7]{SENAI Innovation Institute for Sensing Systems (ISI-SIM), São Leopoldo, RS, Brazil}
\affil[8]{Global Centre for Clean Air Research (GCARE), School of Sustainability, Civil and Environmental Engineering and Physical Sciences, University of Surrey, Guildford, UK}
\affil[9]{Institute for Sustainability, University of Surrey, Guildford, UK}

\date{}

\begin{document}
\maketitle

\begin{abstract}
To enable Industry 5.0, where autonomous systems are widely adopted and integrated with people to promote sustainable development in industries, it is imperative to adapt complex deep learning models to operate efficiently on resource-constrained hardware for edge inference.
However, adapting these models remains challenging due to the hardware heterogeneity of edge computing in industrial settings, performance degradation caused by model compression, and the lack of clear guidance on which compression techniques are practical and effective for applications such as predictive maintenance. To address these challenges, this paper presents a comprehensive comparative analysis of four state-of-the-art model compression techniques applied to a high-performance convolutional neural network for predictive maintenance using a real-world dataset of rotating machinery, equipment typically used in industry. The compressed models were evaluated across a heterogeneous set of nine computing platforms, ranging from highly constrained edge computing devices to general-purpose and high-performance processors, reflecting the diversity and fragmentation typical of industrial edge environments.
By jointly prioritizing classification performance, inference speed, and compression ratio, we found that compression using knowledge distillation emerged as the most effective approach. The model had more than 90\% speedup across all devices while maintaining a 91.22\% balanced F-score, with a very small drop in predictive performance relative to the uncompressed model, along with a 98.9\% size reduction. Through successful deployment across multiple edge platforms and a systematic comparison of compression strategies, we demonstrated a clear path to achieving Industry 5.0 targets, drastically cutting computational costs and, consequently, energy consumption without sacrificing predictive performance.
\end{abstract}


\section{Introduction}
Deep Learning (DL) models involve complex operations that require significant energy, hardware resources, and financial investment. Consequently, their sustainability and environmental impact on an international scale became growing concerns in the field \cite{dhar2020}. Research areas such as Edge Computing, the Internet of Things (IoT), and Tiny Machine Learning (TinyML) play a central role in these discussions, as they propose machine learning--aware architectures, frameworks, techniques, tools and approaches designed for resource-constrained devices at the edge of the network \cite{ray2022}. These technologies enable efficient task execution on hardware with strict limitations, such as devices operating in the milliwatt to watt power range, such as IoT nodes, microcontrollers and mobile devices, while remaining scalable and suitable for large-scale deployments \cite{ray2022}.

With edge computing, it is possible to process data and infer results directly from devices locally, reducing the dependence on continuous network connectivity \cite{li2018}. That technology offers significant advantages across a wide range of applications, with a particularly strong impact in industrial settings and predictive maintenance. By processing data locally in edge devices, it is possible to drastically reduce latency, avoid costly delays, and enable rapid responses that can help prevent fatal incidents or environmental disasters \cite{ray2022}. Moreover, the use of AI in embedded systems has the potential to improve predictive maintenance by continuously monitoring equipment conditions, reducing unplanned downtime, and maintenance costs \cite{rensburg2019}. Furthermore, Edge Computing supports efficient resource utilization by allowing algorithms to be adapted to specific hardware, maximizing performance on constrained platforms such as low-power processors \cite{sze2020}.

Despite its many advantages, the adoption of edge computing systems powered by DL models in industrial settings remains limited. Numerous studies reported strong performance in fault diagnosis using DL techniques \cite{saufi2019}, including the PdM-CNN model \cite{souza2021}, a Convolutional Neural Network (CNN) model that achieved more than 99\% accuracy on the widely used MaFaulDa predictive maintenance dataset \cite{MaFaulDa2025}. However, most of these works overlooked practical deployment aspects, particularly in embedded platforms. In fact, there is a lack of analysis and studies on how predictive maintenance models behave on resource-constrained edge devices \cite{dantas2024}, which has led to several open problems, including:
\begin{itemize}
  \item Lack of understanding and practical examples of how compression techniques impact industry-oriented DL models across heterogeneous hardware platforms for edge computing tasks, i.e., embedded systems with diverse architectures and capabilities.
  \item Limited insights into the potential impact of compression techniques on inference time and accuracy for fault diagnosis models.
  \item Existing DL models often require significant redesign or compression to function efficiently on resource-limited devices \cite{ray2022}. There is a critical knowledge gap in the effective compression of complex, yet powerful models, such as CNNs, for predictive maintenance applications. This is especially true when they are deployed on edge devices in an industrial environment, where the effects of compression strategies on model performance and reliability remain largely unexplored.
\end{itemize}

This research presents a unique analysis of DL model compression techniques applied to a state-of-the-art (SoA) architecture for machinery fault diagnosis across a range of computing platforms. The goal is to advance low-latency and low-memory sustainable AI solutions aligned with Industry 5.0 principles. Through a comparative analysis, we investigated how different devices benefit from compression strategies, assess the feasibility of deploying complex models on heterogeneous hardware, and highlight trade-offs relevant to predictive maintenance applications. The main novel aspects of this work are as follows:
\begin{itemize}
  \item The development of D-PdMCNN, which was shown to be more than 90\% faster across all devices and 98.9\% smaller than the fault detection model with only a 3.79\% drop in the F1-score.
  \item Comprehensive cross-platform benchmarking on a diverse set of computing platforms, spanning microcontrollers, edge devices, and general-purpose/high-performance processors.
  \item A comparative study of multiple, easy to implement, DL compression techniques in a SoA architecture for machinery fault diagnosis.
  \item Comprehensive insights into practical deployment challenges and trade-offs between accuracy, efficiency, and hardware constraints.
\end{itemize}

The remainder of this paper is structured as follows. Section~\ref{sec:methods} details the methodology. Section~\ref{sec:results} presents the experimental results, Section~\ref{sec:discussion} offers a critical discussion of the key findings. Finally, Section~\ref{sec:conclusion} concludes the paper and suggests potential directions for future research.

\section{Methods}
\label{sec:methods}

\begin{figure}[htbp]
\centering
\includegraphics[width=0.55\textwidth]{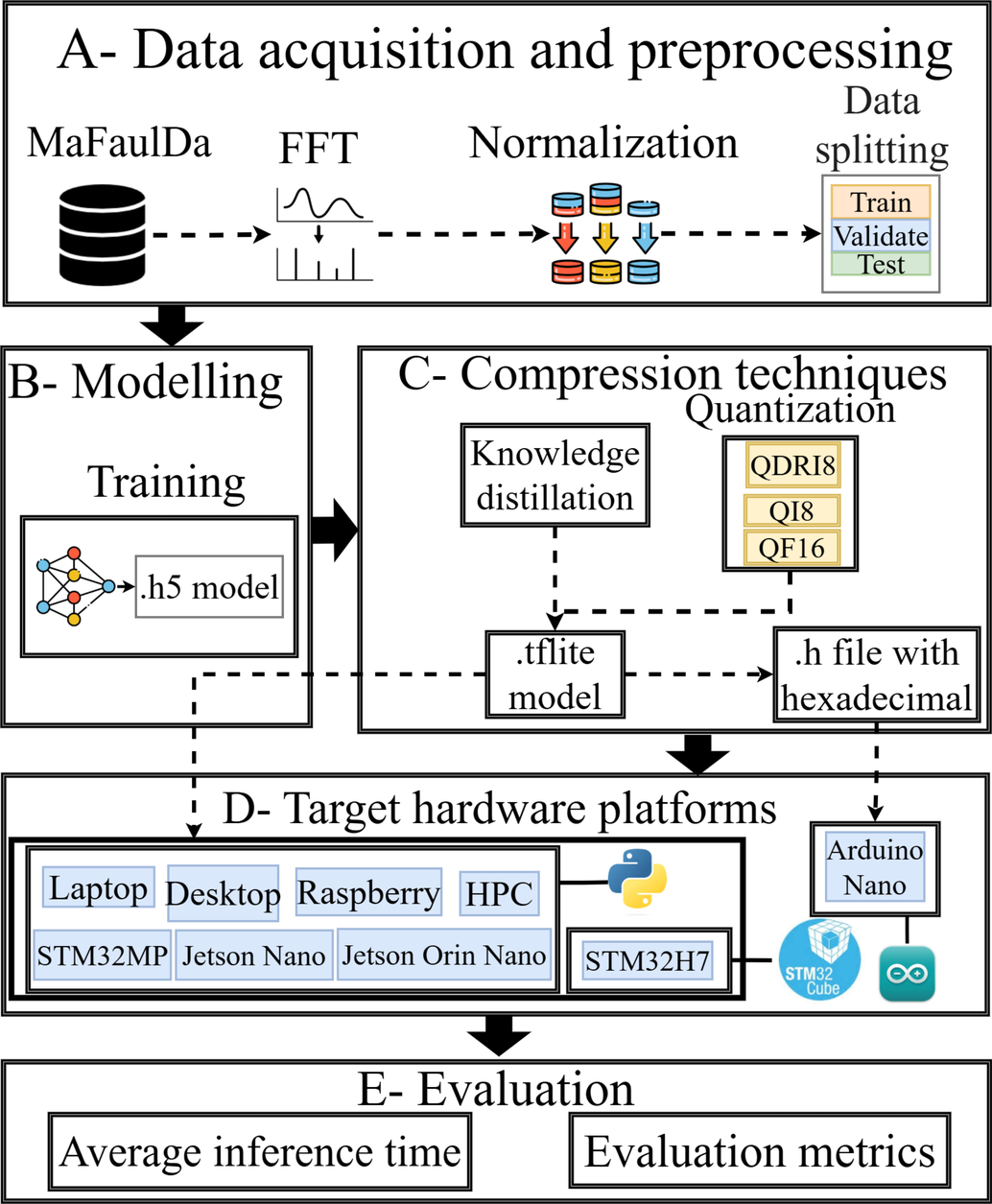}
\caption{Methodological summary for adapting Deep Neural Networks to diverse computing devices.}
\label{fig:pipeline}
\end{figure}

The methodology adopted in this study was structured into five distinct phases, visually depicted in Fig.~\ref{fig:pipeline}. The first phase defines the scope of the data, including a detailed description of the dataset, preprocessing steps, criteria for the definition of the target variables, and the strategy for splitting the data into training, validation, and test sets. The second phase, illustrated in box B of Fig.~\ref{fig:pipeline}, focuses on the modeling process. It outlines the architecture of the model and justifies the changes made in it for this study. The third phase details the applied model compression techniques. This phase also provides key information on how the model can be effectively adapted for deployment across a wide range of hardware platforms. The fourth phase describes the selected computing devices in detail and offers a contextual justification for their inclusion. The final phase introduces the evaluation metrics and performance criteria used to assess the model, explaining their importance in measuring the feasibility and effectiveness of deploying the model on edge devices.

\subsection{Data acquisition and preprocessing}
\begin{figure}[htbp]
\centering
\includegraphics[width=0.8\textwidth]{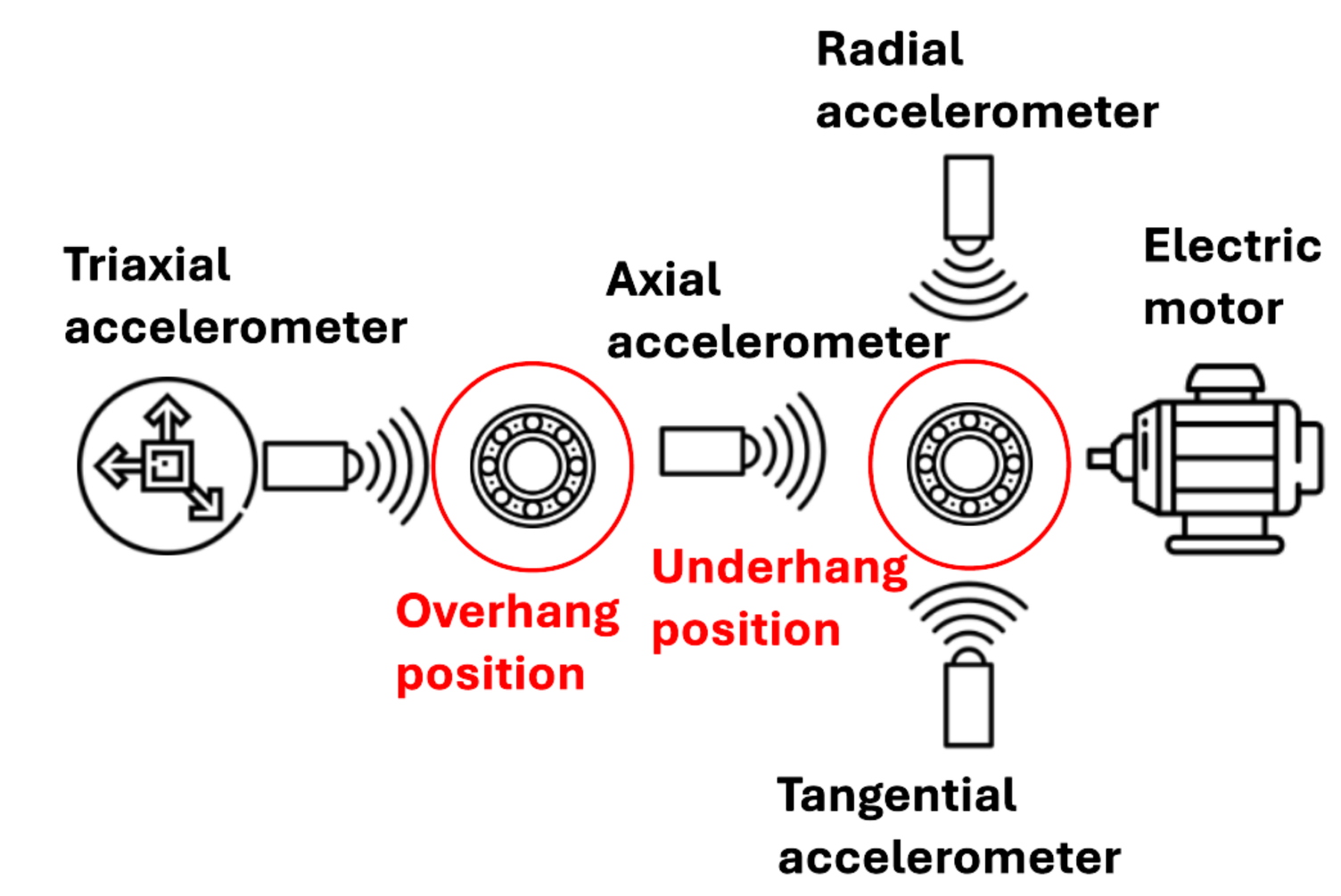}
\caption{MaFaulDa experimental setup (adapted from \cite{MaFaulDa2025}).}
\label{fig:MaFaulDa}
\end{figure}

For benchmarking and training of the baseline model, the Machinery Fault Database (MaFaulDa) \cite{MaFaulDa2025} was used.
The MaFaulDa dataset is a common reference in predictive maintenance projects \cite{saufi2019}. This database contains 1951 multivariate time series collected by sensors on a SpectraQuest Machinery Fault Simulator (MFS) Alignment/Balance Vibration Trainer (ABVT). These 1951 samples have six possible simulated states: normal function, imbalance fault, horizontal and vertical misalignment faults, and inner and outer bearing faults \cite{MaFaulDa2025}. The experimental setup used to collect data from the MaFaulDa dataset is presented in Fig.~\ref{fig:MaFaulDa}.

Data were collected using two sets of accelerometers (102 mV/g sensitivity) positioned longitudinally, horizontally, and vertically, as well as a tachometer for rotor speed and a microphone for noise measurement. Vibration signals were captured at 51.2 kHz using the National Instruments NI 9234 data collection system. The operation conditions \cite{MaFaulDa2025} were the following: (1) \textbf{Normal function} -- error-free operation collected from 49 test situations with rotation speeds ranging from 737 to 3686 rpm. (2) \textbf{Horizontal misalignment} -- motor shaft misalignment obtained at intervals of 0.5, 1.0, 1.5, and 2.0 mm, evaluated over 197 scenarios using normal-operation speeds. (3) \textbf{Vertical misalignment} -- vertical shaft offsets of 0.51, 0.63, 1.27, 1.4, 1.78, and 1.9 mm, resulting in 301 scenarios at normal-operation rotational speeds. (4) \textbf{Imbalance} -- rotor imbalance induced by discrete loads 6 g, 10 g, 15 g, 20 g, 25 g, 30 g, and 35 g, yielding 333 scenarios, with high-load data limited at speeds above 3300 rpm. (5) \textbf{Bearing faults} -- outer race, ball, and cage defects evaluated in underhang and overhang configurations, with added masses of 6, 10, and 20 g to amplify defect signatures in the data. The dataset comprises 558 underhang and 513 overhang scenarios.

The input variables of the baseline model were the signals from three unidirectional accelerometers, which represent a standard monitoring configuration in industrial settings. These accelerometers were located in the underhang position and collect data in the axial, radial, and tangential directions.
We processed these signals under five operational conditions, which were the target variables: normal function, imbalance, horizontal parallel misalignment, bearing cage fault, and bearing outer race fault.
The 5 classes in question were chosen to represent states with a certain variety, thus ensuring a certain degree of adaptability for the model.
To improve the performance of the model, the Fast Fourier Transform (FFT) was implemented in each vibration sensor data, and the resulting amplitudes were used as the input vector for training, validation, and testing.
The feature extraction significantly reduced the size of the input vector compared to the number of samples in a time series. This step also improved the discrimination of failure classes compared to the original time series.
Soon after, the data were normalized using scikit-learn's MinMaxScaler method. A total of 8580 samples were extracted from FFT sequences by cutting them in smaller time windows: 5550 were used for training, 1220 for validation, and 1810 for testing.

\subsection{Modeling}
\begin{figure}[htbp]
\centering
\includegraphics[width=0.85\textwidth]{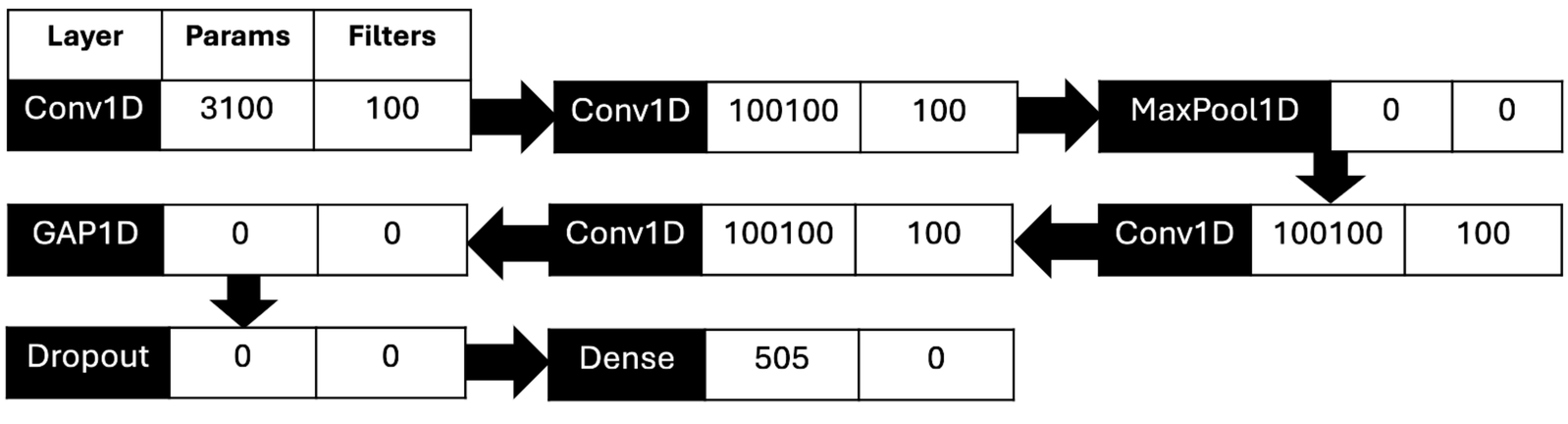}
\caption{PdM-CNN architecture.}
\label{fig:MaFaulDa_architecure}
\end{figure}

The developed model reproduces the PdM-CNN \cite{souza2021} architecture. The main reasons for that are that the model reported strong performance, it used convolutional layers, which have the potential to capture relevant features from data, and it has low hardware requirements since it has 300K parameters, which is capable of fitting in several types of hardware. The model incorporates some main components: the Rectified Linear Unit (ReLU) and the softmax activation functions, one-dimensional convolutional layers (Conv1D), a one-dimensional max pooling layer (MaxPool1D), a one-dimensional global average pooling layer (GAP1D), a dropout layer and a dense layer. The architecture is summarized in Fig.~\ref{fig:MaFaulDa_architecure}. The categorical cross-entropy loss function was utilized during training, and the model parameters were optimized using the Adam optimizer. The model, originally designed for 10 fault classes, was reproduced and adapted to classify 5 classes. This reduction simplified deployment on less powerful devices, making it suitable for edge computing. It also streamlined data acquisition and performance evaluation, allowing for more efficient experimentation across various devices and compression techniques. This model serves as the baseline for all subsequent compression transformations and analytical evaluations.

\subsection{Compression Techniques}
To carry out the study, we subjected the baseline model to four compression techniques: 8-bit Integer Quantization (QI8), Dynamic Range Quantization (QDRI8), Float16 Quantization (QF16), and Knowledge Distillation (KD). These methods were selected for their effectiveness in significantly reducing the size of the model \cite{tensorflow2024}, they are commonly adopted, and easy to reproduce, since they are publicly documented. In addition, their use provides trade-offs between model size, inference latency, hardware compatibility, and accuracy, enabling efficient deployment across CPUs, GPUs, and microcontrollers.

\subsubsection{Post-Training Quantization Methods}
Post-training quantization (PTQ) refers to a class of model compression techniques applied after the completion of training, with the objective of reducing model size, memory bandwidth, and inference latency without modifying the training procedure. PTQ typically operates by reducing the numerical precision of model parameters and, in some cases, intermediate activations. Depending on the target representation, PTQ methods can be broadly categorized as integer quantization and floating-point precision reduction. In this work, all PTQ methods were implemented using the TensorFlow Lite quantization toolchain.

Integer quantization maps real-valued tensors to discrete integer representations. TensorFlow Lite employs an asymmetric affine quantization scheme to convert a real-valued tensor element into a quantized integer value.

Two integer-based PTQ configurations were evaluated:
\begin{itemize}
\item \textbf{Full Integer Quantization (QI8):}
Both weights and activations are statically quantized to 8-bit signed integers (int8). Calibration data are used to estimate fixed scale and zero-point parameters for all quantized tensors, resulting in a fully integer inference pipeline suitable for deployment on integer-only hardware accelerators.

\item \textbf{Dynamic Range Integer Quantization (QDRI8):}
The weights are statically quantized to int8 using affine quantization. Activations are stored in floating-point format and quantized dynamically to 8-bit precision during inference for operations that support integer kernels. After processing, the results are dequantized back to the floating-point format. This approach reduces the size of the model and can improve inference performance while avoiding the need for activation calibration.
\end{itemize}

Floating-point precision reduction decreases numerical precision while preserving floating-point arithmetic semantics. Unlike integer quantization, this approach does not require scale or zero-point parameters and does not involve discrete integer representations.
One floating-point-based PTQ technique was evaluated:
\begin{itemize}
\item \textbf{Float16 Quantization (QF16):}
Model weights are converted from 32-bit floating-point (float32) to 16-bit floating-point (float16) representations. Activations typically remain in float32 during inference, although some hardware accelerators may internally cast computations to float16. This method reduces memory footprint and memory bandwidth while maintaining a wide dynamic range, and is particularly effective on platforms with native float16 support.
\end{itemize}

\subsubsection{Knowledge Distillation}
KD \cite{gou2021} differs from post-training quantization (PTQ) methods in that it requires retraining D-PdMCNN rather than modifying a pretrained network after training. Therefore, KD was evaluated here as a combined compression--retraining strategy rather than a pure post-training transformation. In KD, the knowledge of a large and accurate model (teacher) is transferred to a smaller and more efficient model (student). The key idea is that, instead of training the student solely on ground-truth labels, the student also learns from the teacher's output distribution. This distribution often contains additional information about the relative similarities between classes, which can help the student better generalize. KD is widely used to deploy DL models in resource-constrained environments by reducing model size while preserving accuracy.

In its general form, KD can be performed through different strategies \cite{gou2021}: matching logits (soft-label distillation), aligning feature representations (feature distillation), or transferring attention maps. The most common approach is soft-label distillation, in which the student is trained to mimic the softened output probabilities of the teacher model.

In this work, we apply soft-label distillation. The teacher model, shown in Fig.~\ref{fig:MaFaulDa_architecure}, is a CNN trained on the full dataset, and the student is a lighter CNN with knowledge distilled from the model based on the PdM-CNN architecture. A temperature parameter $T=40$ is applied to the teacher's logits to produce smoother probability distributions:
\begin{equation}
p_i^{(T)} = \frac{\exp(z_i/T)}{\sum_j \exp(z_j/T)}
\label{eq:softmax_temperature}
\end{equation}
where $p_i^{(T)}$ is the softened probability of the teacher model for class $i$ at temperature $T$, $z_i$ is the logit of class $i$ from the teacher model, and $z_j$ is the logit of class $j$ from the teacher model.

Knowledge distillation enables the transfer of information from a larger teacher model to a smaller D-PdMCNN by leveraging softened output probabilities instead of hard labels. Temperature scaling in the softmax function reduces logit dominance and produces smoother probability distributions that expose secondary class probabilities, commonly referred to as dark knowledge \cite{hinton2015distilling}. As the temperature increases, the probability mass is redistributed between classes, altering the relative weighting of teacher predictions without modifying the underlying model parameters.

\begin{figure}[htbp]
\centering
\includegraphics[width=0.85\textwidth]{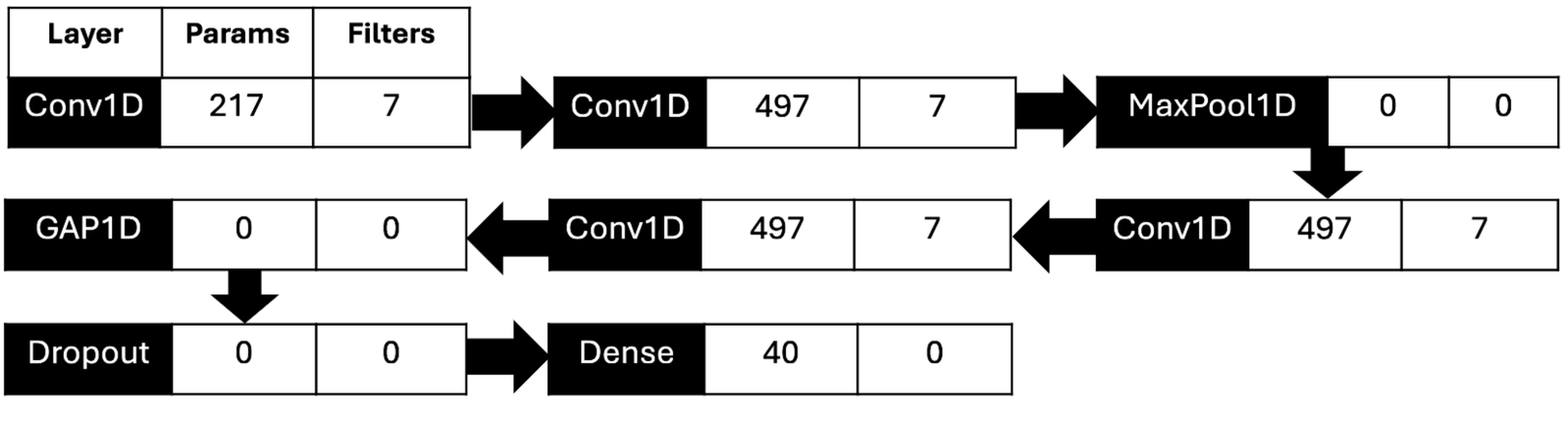}
\caption{Student architecture of the distilled model (D-PdMCNN).}
\label{fig_kd_new}
\end{figure}

The student was optimized using a joint loss that balances ground-truth supervision with teacher guidance:
\begin{equation}
\mathcal{L} = \alpha \cdot H(y,p_s) + (1-\alpha)\cdot T^2 \cdot \mathrm{KL}\!\left(p_t^{(T)} \,\|\, p_s^{(T)}\right)
\label{eq:kd_loss}
\end{equation}
where $\mathcal{L}$ is the total loss used to train D-PdMCNN, $\alpha$ is the weighting factor that balances ground-truth loss and distillation loss ($0 \leq \alpha \leq 1$), $H(\cdot,\cdot)$ is the Sparse Categorical Cross-Entropy between the labels $y$ and the student predictions $p_s$, and $\mathrm{KL}(\cdot\|\cdot)$ is the Kullback--Leibler divergence between the teacher and student softened predictions. In our implementation, $\alpha=0.1$ was used, that is, $90\%$ of the training signal came from the teacher and only $10\%$ came from the ground-truth labels. A low value of $\alpha$ was selected to emphasize teacher guidance, which is particularly important for training very small student networks. This setup prioritized the transfer of knowledge from the teacher's softened predictions to improve generalization.

D-PdMCNN was derived from the baseline model by reducing convolutional filters and fully connected units, producing a compact student network with only 1,748 parameters, see Fig.~\ref{fig_kd_new}.

Finally, it is worth mentioning that for compression, PTQs and KD, all models were converted to the flatbuffer format using TensorFlow Lite, which is a collection of tools that enables on-device machine learning by helping developers run their models on mobile, embedded, and edge devices \cite{tensorflow2024}. The flatbuffer format enables lightweight deployment and efficient runtime execution, particularly when combined with quantization and operator optimizations, enabling TensorFlow Lite to run efficiently on devices with limited compute and memory resources \cite{tensorflow2024}. This allowed the application of these models on multiple devices, as will be explained in the next section.

\subsection{Target Hardware Platforms}
This research evaluated the models on a diverse set of computing platforms, deliberately chosen to represent a broad spectrum of hardware architectures and computing capabilities, including variations in frequency. This hardware heterogeneity mirrored the practical deployment challenges in industrial environments. Evaluating the performance of machine learning models on a variety of development platforms is essential to determine their viability and effectiveness in real-world applications, especially in resource-constrained environments. The platforms tested range from general-purpose processors (e.g., desktop with Intel\textregistered{} Core\texttrademark{} i5-10400F, laptop with Intel\textregistered{} Core\texttrademark{} i5-11300H, and HPC node with Intel\textregistered{} Xeon\textregistered{} Gold 6148), to microcontrollers (e.g., Arduino Nano BLE Sense 33 and STM32H7A3ZIT6Q), and specialized edge devices (e.g., Jetson Nano, Jetson Orin Nano, Raspberry Pi 4B, and STM32MP157F-DK2), covering a substantial portion of the microcontrollers and edge devices available on the market. For some devices, we also explored the use of multiple CPU cores (MCPU) at the same time. Detailed specifications of the devices are presented in Table~\ref{tab:devices_description}. This approach ensures that embedded AI systems meet key requirements such as performance, accuracy, and reliability within their specific industrial contexts.

In practice, adapting DL to all hardware in Table~\ref{tab:devices_description} was challenging. For our experiments, we ran inference on each of the 1810 samples of the test set using the baseline model and the versions resulting from the various compression techniques. Python code handled these benchmarks for all devices except Arduino Nano BLE Sense 33 and STM32H7A3ZIT6Q. STM32CubeIDE (an integrated development environment) was used for inference on the STM32H7A3ZIT6Q microcontroller. For Arduino, the model was converted to a hexadecimal array within an \texttt{.h} file, which was then used in the Arduino code through the Arduino IDE. As TensorFlow Lite does not natively support CUDA-based NVIDIA GPUs (e.g., Jetson platforms), the models in TFLite format could not be executed using GPU acceleration. In this inference mode on Jetson Nano and Jetson Orin Nano, we ran only the baseline model in \texttt{.h5} format.

\begin{table}[htbp]
\caption{Devices used for the comparative study of DL model compression techniques.}
\label{tab:devices_description}
\centering
\small
\setlength{\tabcolsep}{4pt}
\begin{tabular}{@{}llcllr@{}}
\toprule
\textbf{Device} & \textbf{Processor} & \textbf{Cores} & \textbf{Architecture} & \textbf{Frequency} & \textbf{Memory} \\
\midrule
Desktop & Intel Core i5-10400F & 6 & x86-64 & 2.90 GHz & 16 GB \\
Laptop & Intel Core i5-11300H & 4 & x86-64 & 2.60 GHz & 8 GB \\
HPC & Intel Xeon Gold 6148 & 20 & x86-64 & 2.40 GHz & 768 GB \\
STM32M & Dual Arm Cortex-A7 & 2 & ARMv7-A & 800 MHz & 512 MB \\
Raspberry & Quad-core ARM Cortex-A72 & 4 & ARMv8-A & 1.5 GHz & 8 GB \\
J. Nano & Quad-core ARM Cortex-A57 & 4 & ARMv8-A & 1.43 GHz & 4 GB \\
 & NVIDIA Maxwell 128-core (GPU) & 128 & NVIDIA Maxwell & 1.43 GHz & 4 GB \\
J. Orin Nano & ARM Cortex-A78AE v8.2 6-core & 6 & ARMv8.2-A & 1.5 GHz & 8 GB \\
 & GA10B (GPU) & 1024 & NVIDIA Ampere & 625 MHz & 8 GB \\
Arduino & ARM Cortex-M4 & 1 & ARMv7E-M & 64 MHz & 256 KB \\
STM32H & ARM Cortex-M7 & 1 & ARMv7E-M & 280 MHz & 1.4 MB \\
\bottomrule
\end{tabular}
\end{table}

\subsection{Evaluation}
Evaluating the performance of machine learning models across diverse development platforms is a critical step in assessing their suitability for real-world applications, particularly in environments with limited computational resources. In the context of embedded AI systems, this evaluation ensures that the models meet essential operational criteria such as execution performance, predictive accuracy, and system reliability. These factors are especially significant in industrial applications, where deployment conditions can vary significantly and demand high resilience.
Compressing models to fit low-power devices often leads to a reduction in performance. Therefore, when evaluating results, it is important to determine the extent of this performance loss and determine its practical impact on predictive maintenance applications.
The performance of the model was evaluated using five widely used classification metrics: precision, recall, F1-score, specificity, and accuracy.
To provide a view of the models' capabilities, we evaluated not only performance metrics but also the average inference time on each device. In addition, the sizes of the models in flatbuffer format were assessed for each compression method.

\FloatBarrier
\section{Results}
\label{sec:results}

\begin{figure}[htbp]
\centering
\includegraphics[width=0.6\textwidth]{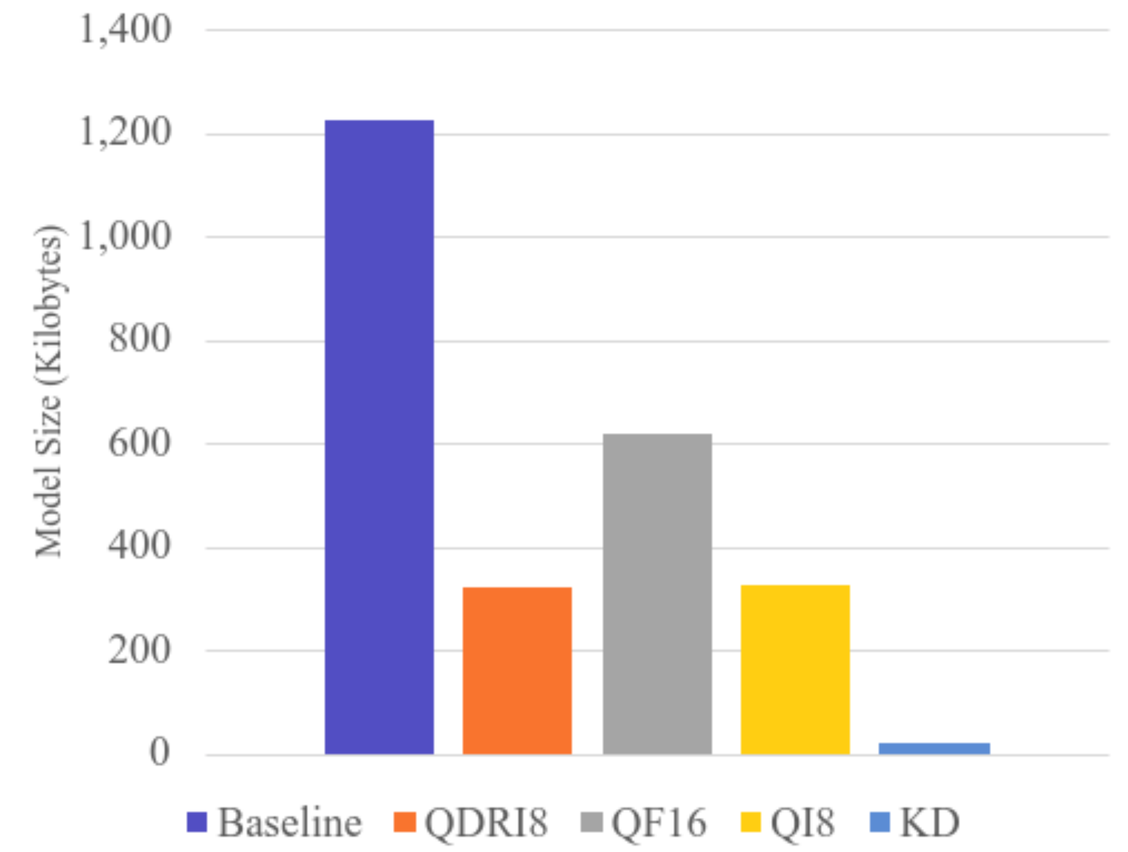}
\caption{Bar plot presenting the model size reduction achieved across different compression techniques relative to the baseline.}
\label{figsize}
\end{figure}

Figure~\ref{figsize} summarizes the reduction in memory footprint achieved by the evaluated compression techniques. As expected, more aggressive optimization strategies provide greater storage savings, making the resulting models increasingly suitable for deployment on embedded and edge devices with limited memory resources. However, memory footprint alone is not sufficient to determine the practical value of a compressed model. In real-world applications, the reduction in model size must also translate into efficient execution without introducing excessive computational overhead.

\begin{table}[htbp]
\caption{Average inference time (ms) for each device under the different compression techniques. $\uparrow$ indicates percentage growth and $\downarrow$ percentage reduction relative to the baseline.}
\label{tab:inference_times}
\centering
\scriptsize
\resizebox{\textwidth}{!}{%
\begin{tabular}{@{}llrrrrr@{}}
\toprule
\textbf{Device} & \textbf{Mode} & \textbf{Baseline} & \textbf{QDRI8} & \textbf{QF16} & \textbf{QI8} & \textbf{KD} \\
\midrule
Desktop & CPU & 14.69 & 17.33 $\uparrow$(18.0\%) & 15.13 $\uparrow$(3.0\%) & 7.99 $\downarrow$(45.6\%) & 0.34 $\downarrow$(97.7\%) \\
Laptop & CPU & 7.77 & 10.98 $\uparrow$(41.4\%) & 7.60 $\downarrow$(2.1\%) & 5.81 $\downarrow$(25.2\%) & 0.20 $\downarrow$(97.5\%) \\
HPC & CPU & 7.34 & 12.44 $\uparrow$(69.5\%) & 7.38 $\uparrow$(0.6\%) & 7.47 $\uparrow$(1.7\%) & 0.34 $\downarrow$(95.3\%) \\
STM32M & CPU & 784.95 & 499.38 $\downarrow$(36.4\%) & 772.23 $\uparrow$(1.6\%) & 393.08 $\downarrow$(49.9\%) & 9.96 $\downarrow$(98.7\%) \\
Raspberry & CPU & 81.77 & 57.22 $\downarrow$(30.0\%) & 83.68 $\uparrow$(2.3\%) & 54.97 $\downarrow$(32.8\%) & 1.85 $\downarrow$(97.7\%) \\
\midrule
\multirow{3}{*}{J. Nano} & CPU & 148.52 & 98.55 $\downarrow$(33.6\%) & 150.65 $\uparrow$(1.4\%) & 93.05 $\downarrow$(37.4\%) & 2.58 $\downarrow$(98.3\%) \\
 & MCPU & 76.45 & 62.91 $\downarrow$(17.7\%) & 77.35 $\uparrow$(1.2\%) & 57.66 $\downarrow$(24.6\%) & 2.25 $\downarrow$(97.1\%) \\
 & GPU & 52.16 & -- & -- & -- & -- \\
\midrule
\multirow{3}{*}{J. Orin Nano} & CPU & 38.11 & 13.64 $\downarrow$(64.2\%) & 38.79 $\uparrow$(1.8\%) & 39.52 $\uparrow$(3.7\%) & 0.23 $\downarrow$(99.4\%) \\
 & MCPU & 8.50 & 11.92 $\uparrow$(40.2\%) & 9.10 $\uparrow$(7.1\%) & 15.93 $\uparrow$(87.4\%) & 0.80 $\downarrow$(90.6\%) \\
 & GPU & 5.52 & -- & -- & -- & -- \\
\midrule
Arduino & CPU & -- & -- & -- & -- & 1096.89 \\
STM32H & CPU & -- & -- & -- & -- & 52.13 \\
\bottomrule
\end{tabular}}
\end{table}

Table~\ref{tab:inference_times} presents the average inference time measured on the target hardware devices. The results demonstrate how each compression technique affects computational efficiency, revealing the trade-off between reducing the model footprint and achieving faster execution. Since embedded AI applications typically require both low latency and high predictive reliability, inference time should not be analyzed independently. A compression method that significantly accelerates execution but substantially degrades classification performance may not be appropriate for practical deployment.

\begin{table}[htbp]
\caption{Comparative performance metrics of models across multiple hardware devices. Results were identical across all devices (Desktop, Laptop, HPC, STM32M, Raspberry, J.~Nano, J.~Orin Nano, Arduino and STM32H); Arduino and STM32H ran only the KD model. $^*$Baseline model has no compression. $^{**}$D-PdMCNN trained from scratch with randomly initialized weights. Prec.\ = precision, Spec.\ = specificity, Acc.\ = accuracy.}
\label{tab:performance_metrics}
\centering
\small
\resizebox{\textwidth}{!}{%
\begin{tabular}{@{}lccccc@{}}
\toprule
\textbf{Model} & \textbf{F1} & \textbf{Prec.} & \textbf{Recall} & \textbf{Spec.} & \textbf{Acc.} \\
\midrule
PdM-CNN$^*$ & 0.9481 & 0.9590 & 0.9391 & 0.9903 & 0.9852 \\
QDRI8 & 0.9485 $\uparrow$ (0.04\%) & 0.9577 $\downarrow$ (0.13\%) & 0.9406 $\uparrow$ (0.16\%) & 0.9903 & 0.9852 \\
QF16 & 0.9481 & 0.9590 & 0.9391 & 0.9903 & 0.9852 \\
QI8 & 0.9495 $\uparrow$ (0.15\%) & 0.9612 $\uparrow$ (0.23\%) & 0.9401 $\uparrow$ (0.11\%) & 0.9906 $\uparrow$ (0.03\%) & 0.9856 $\uparrow$ (0.04\%) \\
D-PdMCNN (KD) & 0.9122 $\downarrow$ (3.79\%) & 0.9176 $\downarrow$ (4.32\%) & 0.9102 $\downarrow$ (3.08\%) & 0.9822 $\downarrow$ (0.82\%) & 0.9717 $\downarrow$ (1.37\%) \\
D-PdMCNN$^{**}$ & 0.7298 $\downarrow$ (23.03\%) & 0.8328 $\downarrow$ (13.16\%) & 0.7313 $\downarrow$ (22.12\%) & 0.9588 $\downarrow$ (3.18\%) & 0.9364 $\downarrow$ (4.95\%) \\
\bottomrule
\end{tabular}}
\end{table}

Table~\ref{tab:performance_metrics} reports the classification performance achieved after applying each compression strategy. Together with the latency measurements presented previously, these results provide a comprehensive assessment of the effectiveness of the proposed framework, allowing the identification of compression techniques that simultaneously preserve predictive accuracy, reduce execution time, and minimize memory consumption. Among the evaluated approaches, D-PdMCNN consistently achieves a favorable balance between these competing objectives, demonstrating its suitability for deployment on heterogeneous edge hardware.

\begin{table}[htbp]
\caption{\textbf{Comparison with related works:} hardware diversity, compression technique coverage, and the trade-off between performance and compression rate achieved by the proposed model, D-PdMCNN. \#Dev = number of devices; MCU = microcontroller; SBC (Acc.\ Edge) = single-board computers (number of which are accelerated edge devices); Proc = consumer and high-performance processors; \#Tech = number of techniques; KD = knowledge distillation; QI8 = 8-bit integer quantization; QF16 = Float16 quantization; QDRI8 = dynamic range quantization; Prun = weight pruning; Clust = weight clustering; Compr\% = compression rate; PerfLoss\% = performance loss vs.\ baseline; Acc = accuracy. $^{\wedge}$Used hard sigmoid activation functions. $^{\wedge\wedge}$Used depthwise separable convolution.}
\label{tab:related_works}
\centering
\scriptsize
\resizebox{\textwidth}{!}{%
\begin{tabular}{@{}r*{14}{c}@{}}
\toprule
 & \multicolumn{4}{c}{\textbf{Hardware Heterogeneity}} & \multicolumn{7}{c}{\textbf{Compression Techniques}} & \multicolumn{3}{c}{\textbf{Models and Performance}} \\
\cmidrule(lr){2-5}\cmidrule(lr){6-12}\cmidrule(lr){13-15}
\textbf{Ref.} & \textbf{\#Dev} & \textbf{MCU} & \textbf{SBC (Acc.)} & \textbf{Proc} & \textbf{\#Tech} & \textbf{KD} & \textbf{QI8} & \textbf{QF16} & \textbf{QDRI8} & \textbf{Prun} & \textbf{Clust} & \textbf{Compr\%} & \textbf{PerfLoss\%} & \textbf{DL Model(s)} \\
\midrule
\cite{chen2023} & 6 & \textbf{4} & 1(0) & 1 & 3$^{\wedge}$ & -- & \checkmark & -- & -- & \checkmark & -- & 75--87.5\% & 5--24\% acc & 2 LSTMs \\
\cite{rachmanto2024} & 2 & 2 & 0(0) & 0 & 1 & -- & -- & -- & -- & \checkmark & -- & 55.5--99.7\% & 0.72--3.25\% acc & 6 models \\
\cite{mou2024} & 2 & 1 & 1(0) & 0 & 1 & -- & -- & -- & -- & -- & \checkmark & 61.2--85.7\% & 0--89.8\% acc & 7 models \\
\cite{dhar2024} & 2 & 0 & 1(1) & 1 & 1 & \checkmark & -- & -- & -- & -- & -- & 96.6\% & 0.44--1.61\% acc & 6 models \\
\cite{chen2021} & 1 & 1 & 0(0) & 0 & 3 & -- & \checkmark & -- & -- & \checkmark & \checkmark & 14.2--74.6\% & 2.4--16.6\% RMSE & LSTM, CNN \\
\cite{pandey2023deploying} & 1 & 1 & 0(0) & 0 & 3 & -- & \checkmark & -- & -- & \checkmark & \checkmark & 0.4--50.4\% & 0--1.45\% acc & 3 models \\
\cite{an2023} & 1 & 0 & 1(0) & 0 & 1 & -- & -- & -- & -- & \checkmark & -- & 65.8\% & 0.26\% acc & 6 models \\
\cite{wang2024} & 1 & 0 & 1(0) & 0 & 2$^{\wedge\wedge}$ & \checkmark & -- & -- & -- & -- & -- & 77.3--91.8\% & 0--3\% acc & ResNet, CNN \\
\cite{athanasakis2022} & 2 & 2 & 0(0) & 0 & 3 & -- & \checkmark & -- & -- & \checkmark & -- & 23.1--91.6\% & 0\% acc & AE, CNN \\
\midrule
\textbf{Ours} & \textbf{9} & 2 & \textbf{4(2)} & \textbf{3} & \textbf{4} & \checkmark & \checkmark & \checkmark & \checkmark & -- & -- & 50--98.9\% & 0--1.37\% acc & CNN, D-PdMCNN \\
\bottomrule
\end{tabular}}
\end{table}

\FloatBarrier
\section{Discussion}
\label{sec:discussion}
By analyzing Table~\ref{tab:inference_times} of the average inference times for different compression techniques on various devices, one can see that QF16 performed worse than baseline, i.e., the one with no compression, in almost all platforms, as indicated by the upward arrows in the QF16 column. Similarly, QDRI8 and QF16 showed slightly higher inference times compared to the baseline model for the Desktop device. This suggests that, although these techniques can reduce the model size, they can increase the time required to make inferences depending on the environment. On the other hand, for the laptop device, QF16, QI8 and KD produced lower inference times than the baseline model, indicating an improvement in inference performance. In fact, the table highlights notable volatility in the percentage of speed reduction or improvement relative to the baseline, particularly for QDRI8 and QI8. For example, QDRI8 was 69.49\% slower than the baseline on the HPC, yet 64.22\% faster on the Jetson Orin Nano (CPU). Similarly, QI8 was 87.41\% slower on the Jetson Orin Nano (MCPU), but achieved a 49.92\% speedup on the STM32MP157F-DK2.

Despite volatility, some compression techniques can reduce the inference time relative to the baseline model, depending on the target hardware and the numerical format. However, the magnitude of this reduction varies depending on the compression technique and device used. Furthermore, the only technique that consistently reduced the average inference time across all devices was KD, as shown in the last column of Table~\ref{tab:inference_times}; the proposed approach not only consistently reduces the inference time, but does so significantly. It achieved a 99.38\% reduction in inference time on the Jetson Orin Nano (CPU). KD consistently reduces inference time because it results in a much smaller architecture, rather than due to changes in numerical precision. Furthermore, it should be noted that even highly resource-constrained devices that were unable to execute the largest models, such as the STM32H7A3ZIT6Q and Arduino Nano BLE Sense 33, could execute the D-PdMCNN. In particular, the STM32H7A3ZIT6Q achieved an average inference time of 52.13 ms, which was faster than the uncompressed model running on more powerful embedded platforms such as the Jetson Nano (CPU), which took 148.52 ms.

When analyzing Fig.~\ref{figsize} of the sizes of the model submitted to different techniques in the flatbuffer format, we observed a significant reduction in the size of the models after applying the compression techniques. All reduced models had considerably smaller sizes compared to the baseline model, which was 1,222.696 kilobytes. The reduction in size varied significantly between reduction techniques. For example, the QDRI8 model was 318.616 kilobytes, whereas the QF16 model was 616.348 kilobytes, showing that the QF16 technique results in a larger model size compared to QDRI8. The KD model was the smallest, having only 13.520 kilobytes, which corresponds to a substantial reduction in size of approximately 98.9\% compared to the baseline.

On Table~\ref{tab:performance_metrics} one can see the performance metrics for the model without compression on multiple devices, as well as its compressed versions on multiple platforms. The first line shows the results of the baseline model for different metrics, and the remaining lines show the result of the compression technique, deployed in multiple devices (Arduino Nano BLE Sense 33 and STM32H7A3ZIT6Q were executed only using the KD technique due to hardware constraints). The arrows indicate the gain/loss of performance compared to the baseline. A third model with the same architecture as D-PdMCNN was trained from scratch with the same data to have a real sense of the benefits of using soft-label knowledge distillation. Analyzing the data in Table~\ref{tab:performance_metrics}, we observed that the metrics for the different methods (QDRI8, QF16, QI8, KD) were close, suggesting consistency in the results. For example, there was no drop of accuracy for QDRI8 and QF16 and only 1.37\% for the KD technique. The biggest change was a decrease of 4.32\% in precision for KD. Although there were small variations between methods, as seen in the precision and F1 metric, these variations can be insignificant depending on the project context. In our experiments, performance metrics remained consistent across devices, indicating that compression techniques, rather than hardware differences, were the dominant factor; this is important because it means we can implement our findings into a large number of different devices. All metrics (F1, precision, recall, specificity, and accuracy) present high results, indicating robust performance across all methods. Precision and accuracy were also consistently above 90\%, suggesting reliable and consistent results even for a model as small as the model resulting from KD that manages to maintain considerably good metrics across all devices. The differences between D-PdMCNN and its version trained from scratch in terms of F1-score and recall were greater than 19\%, demonstrating that the use of knowledge distillation was important to preserve performance.

The underlying reasons for the observed reduction in model size, while maintaining strong evaluation metrics, remain unclear for quantization techniques such as QDRI8 and QF16. This behavior is likely related to the robustness of convolutional feature extractors to reduced numerical precision.
In particular, for KD, it can be argued that the effects happened because D-PdMCNN successfully learned from the baseline model, with its parameters being effectively optimized. As a result, the compressed model maintained strong performance on fault diagnosis tasks from the MaFaulDa dataset, despite having a significantly smaller architecture.

Finally, Table~\ref{tab:related_works} summarizes the impacts of our work in the scientific community. Among the key highlights was the hardware heterogeneity, which showed that our work involved more equipment in number and a broader diversity of edge devices, alongside general-purpose and high-performance processors. In addition, the findings of this study provide compelling evidence that KD was the most effective compression technique among those evaluated when compared with other techniques. The creation of a D-PdMCNN model with performance loss levels between 0 and 1.37\% and with a compression rate of 98.9\% was also another point that stood out compared to related works.

\section{Conclusion}
\label{sec:conclusion}
By analyzing the metrics, reliable performance can be seen in a diverse set of devices with different computing architectures, processing capability, and memory capacity. The reduction in model size varies significantly between techniques, indicating that different methods produce different compression results, which may be interesting for devices with low storage memory. When choosing the compression technique, one must take into account not only the reduction in model size but also the impact on inference time, especially on less robust processors.

This research highlights the direct trade-offs between CNN compression and performance across various devices. KD was shown to be the most effective compression technique among those evaluated. It achieved the smallest model size and the shortest average inference time while maintaining performance metrics close to the original baseline model. It was also demonstrated that using a model with the same architecture as D-PdMCNN but trained from scratch would not be ideal, as important metrics such as F1 and recall presented performance loss greater than 20\% in comparison to the baseline.

To explain why model compression maintains high performance, future research should investigate alternative strategies such as pruning or hybrid methodologies. Assessing generalizability will also require testing across more diverse datasets and architectures. Although direct energy consumption measurements were outside the scope of this study, inference latency and model size are widely used proxies for deployment feasibility on resource-constrained edge devices; future studies should assess that.

\section*{Acknowledgements}
We thank the National Council for Scientific and Technological Development (CNPq) for partially funding this work through the CNPq Universal project 405706/2021-9. We also thank the Supercomputing Center for Industrial Innovation (CS2i) from SENAI CIMATEC, for providing HPC infrastructure support, and the Surrey Institute for People-Centred AI from the University of Surrey (UK) for their institutional support. Erick G. Sperandio Nascimento is a CNPq technological development fellow (Proc. 313527/2026-1). Prashant Kumar acknowledges the support received through the UKRI-funded GREENIN Micro Network Plus (APP55977), RECLAIM Network Plus (EP/W034034/1), GP4Streets (UKRI1281), GreenCities (NE/X002799/1), and UGPN-funded GREENICON projects.

\end{document}